\pdfoutput=1

\documentclass[11pt]{article}

\usepackage[]{acl}

\usepackage{times}
\usepackage{latexsym}

\usepackage[T1]{fontenc}

\usepackage[utf8]{inputenc}

\usepackage{microtype}

\usepackage{graphicx}
\usepackage{multirow}
\usepackage{booktabs}
\usepackage{amssymb}
\usepackage{subfigure}
\usepackage{bbding}
\usepackage{pifont}
\usepackage{amsmath}

\title{DialogVED: A Pre-trained Latent Variable Encoder-Decoder Model for Dialog Response Generation}

\author{Wei Chen\textsuperscript{\rm 1}\thanks{~~Worked during the internship at Microsoft Research Asia. Zhongyu Wei and Yeyun Gong are corresponding authors.}, Yeyun Gong\textsuperscript{\rm 2}, Song Wang\textsuperscript{\rm 3}, Bolun Yao\textsuperscript{\rm 4}, Weizhen Qi\textsuperscript{\rm 2}, Zhongyu Wei\textsuperscript{\rm 1}, \\ \bf Xiaowu Hu\textsuperscript{\rm 2}, Bartuer Zhou\textsuperscript{\rm 2}, Yi Mao\textsuperscript{\rm 3}, Weizhu Chen\textsuperscript{\rm 3}, Biao Cheng\textsuperscript{\rm 2} and Nan Duan\textsuperscript{\rm 2} \\ \textsuperscript{\rm 1}School of Data Science, Fudan University, China \textsuperscript{\rm 2}Microsoft Research Asia, China \\ \textsuperscript{\rm 3}Microsoft, US \textsuperscript{\rm 4}Nanjing University of Science and Technology, China \\ 
\{chenwei18,zywei\}@fudan.edu.cn; \\ \{yegong,weizhen,xiaowuhu,bazhou,bicheng,nanduan\}@microsoft.com; \\
\{sonwang,yimao,wzchen\}@microsoft.com; yaobl001@njust.edu.cn}

\begin{document}
\maketitle
\begin{abstract}
Dialog response generation in open domain is an important research topic where the main challenge is to generate relevant and diverse responses. In this paper, we propose a new dialog pre-training framework called DialogVED, which introduces continuous latent variables into the enhanced encoder-decoder pre-training framework to increase the relevance and diversity of responses. With the help of a large dialog corpus (Reddit), we pre-train the model using the following 4 tasks adopted in language models (LMs) and variational autoencoders (VAEs): 1) masked language model; 2) response generation; 3) bag-of-words prediction; and 4) KL divergence reduction. We also add additional parameters to model the turn structure in dialogs to improve the performance of the pre-trained model. We conduct experiments on PersonaChat, DailyDialog, and DSTC7-AVSD benchmarks for response generation. Experimental results show that our model achieves the new state-of-the-art results on all these datasets. 
\end{abstract}

\section{Introduction}



Pre-trained language models (PLMs) have been widely explored both in natural language understanding (NLU) and generation (NLG) in recent years, this pre-training and fine-tuning paradigm sheds light on various downstream tasks in natural language processing (NLP). Compared with general pre-trained models, task-oriented pre-trained models (such as \emph{Summarization}, \emph{Dialog} and etc.), which is designed in line with task characteristics, may achieve better performance and be more robust. In this paper, we proposes a novel pre-trained dialog response generation model based on previous research.


Dialogue Response Generation (DSG) in open domain is a challenging task with a wide range of application scenarios. Recent advances in DSG utilize pre-trained language models (PLMs) such as BERT \cite{devlin2019bert} and GPT2 \cite{radford2019language} in two major categories. The first one focuses on how to fine-tune PLMs in downstream tasks and address the various application-specific needs and challenges \cite{lin2020exploring}. The second one augments dialog specific tasks into the PLM training \cite{zhang2019dialogpt,bao-etal-2020-plato} and then fine-tunes the new pre-trained model in downstream tasks. We study the latter in this paper.

There is a proverbial one-to-many problem in DSG, i.e., a single dialog context could be followed by multiple reasonable responses. Existing works introduce latent variables to model this problem. For example, VHRED \cite{serban2017hierarchical} incorporates latent continuous variable into the sequence-to-sequence (Seq2Seq) RNN model to improve the diversity of generated responses. VAE-Seq2Seq \cite{bahuleyan2017variational} proposes variational attention to replace the vanilla encoder-decoder attention \cite{luong2015effective}, to avoid attention to bypass the latent space and invalidate the latent variable. For controllability and interpretability, some discrete VAEs have also been proposed, such as~\cite{oord2017neural, vahdat2018dvae++}.

Recently, PLATO \cite{bao-etal-2020-plato} firstly introduces latent variables into their pre-training dialog model, where the authors introduce a $K$-way ($K=20$) categorical latent variable, and the pre-trained model shows significant gains in multiple downstream response generation tasks. Continuous latent variables besides discrete latent variables is popularly used for modeling one-to-many mapping in dialog system, but the potential of incorporating continuous latent variables with large-scale language pretraining is less explored.  


In this paper, we propose a pre-trained latent \textbf{V}ariable \textbf{E}ncoder-\textbf{D}ecoder model for \textbf{Dialog} generation, which is called DialogVED. In this model, we introduce a continuous latent variable into the enhanced encoder-decoder pre-training framework and we adopt the optimization techniques based on the VAEs literature to learn the model with continuous latent variables. More specifically, we conduct the pre-training by optimizing the following 4 pre-training objectives simultaneously: 1) masked language spans loss to enhance the encoder's understanding of context, 2) response generation with n-gram loss to improve the decoder's planning ability, 3) Kullback-Leibler divergence loss to minimize the difference between the posterior and prior distribution of the latent variables, and 4) bag-of-words loss to reduce posterior distribution collapse. In addition, we also explore the effect of absolute and relative position embeddings specific for conversational data on the model performance. 

We conduct experiments on three different kinds of conversation tasks: chit-chat, knowledge grounded conversation, and conversational question answering. Experimental results verify the effectiveness and superiority of our model compared with the previous state-of-the-art method. We further carry out ablation study to better understand the impact of different components in the DialogVED on model performance including latent space sizes, different decoding strategies, and position embeddings for turns and roles. 


The main \textbf{contributions} of this paper can be summarized as follows: 1) We propose a pretrained dialog model, which incorporates continuous latent variables into the enhanced encoder-decoder pre-training framework; 2) We explore the impact of latent variable sizes, different decoding strategies, and position embeddings for turns and roles in our model; 3) Extensive experiments show that the proposed model achieves the new state-of-the-art (SOTA) in multiple downstream tasks, and our model has better performance both on relevance and diversity than previous SOTA in response generation.



\section{Method}
\label{section:method}



\begin{figure*}[htbp]
\centering{
\includegraphics[width=0.9\textwidth]{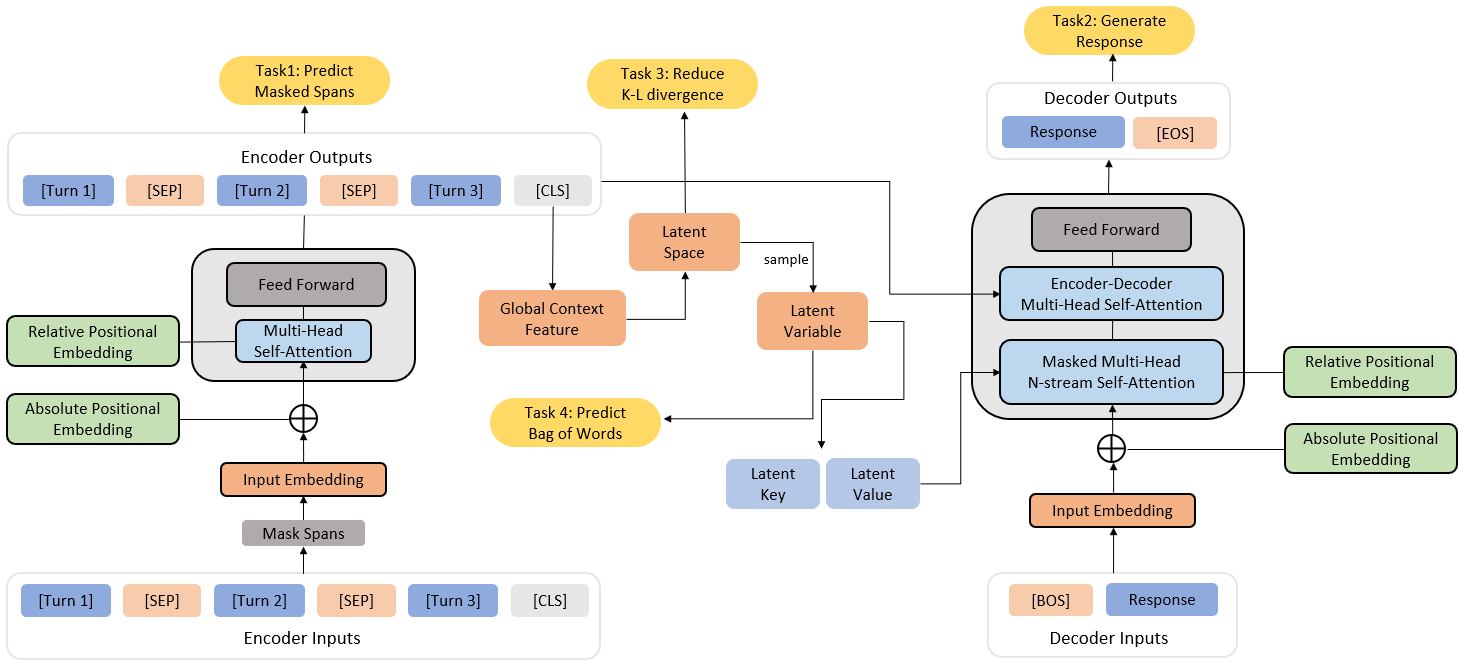}
}
\caption{Pre-training and fine-tuning framework of DialogVED, the only difference between pre-training and fine-tuning is that in the fine-tuning stage, we do not mask the source thus the masked spans loss is discarded. It’s worth noting that, to facilitate drawing, we put [CLS] at the end of the context, although we actually put it at the beginning.}
\label{model}
\end{figure*}


\subsection{Model Architecture}

In response generation, there are three elements: dialogue context $c$, response $r$ and latent variable $z$. The dialogue context $c$ may consist of several history utterances (i.e., multi turns) and the response $r$ is one piece of appropriate reply towards the given context. Additionally, the latent variable $z$ in the latent space represents many unobserved factors associating the context and the response. 


We assume the latent variable $z$ is continuous, which is different from PLATO \cite{bao-etal-2020-plato}, and portrays a certain conditional probability distribution related to the response given context. We then define the conditional distribution $p(r,z|c) = p(r|c,z) p(z|c)$ and our goal is to use encoder-decoder models (parameterized by $\theta$) to approximate $p(r|c,z)$ and a multi-layer perceptron (parametrized by $\phi$) to estimate $p(z|c)$, which is called the prior network in VAE literature. We call the final pre-trained model DialogVED, which is a transformer-based encoder-decoder model with an extra prior network for modeling the latent space. Figure \ref{model} gives a overview of our model.


\subsection{Encoder}

We use multi-layer Transformer-based \cite{vaswani2017attention} encoder to encode the dialogue context. First, an input sequence of tokens is mapped to a sequence of embeddings, which are then passed into the encoder. The encoder consists of a stack of “blocks”, each of which comprises two subcomponents: a self-attention layer followed by a small feed-forward network. Compared to the vanilla transformer encoder, our encoder has slight differences in position embeddings and self-attention layer in fine-tuning phase, which contains richer location information and will be introduced in \S~\ref{section:pe}.

\subsection{Decoder}

Future predicting strategy has been concerned in recent research \cite{qi2020prophetnet, xiao2020ernie}, instead of predicting only the next token at each time step, the decoder using future predicting predicts $n$ future tokens simultaneously. 

Specifically, the original Seq2Seq model aims to optimize the conditional likelihood $P(r_{t}|r_{<t}, c)$, while future predicting strategy changes the optimization of predicting next single token to $P(r_{t:t+n-1}|r_{<t}, c)$ at each time step $t$, where $r_{t:t+n-1}$ denotes the next continuous $n$ future tokens. The future n-gram prediction loss can explicitly encourage the model to plan for future token prediction and prevent over-fitting on strong local correlations \cite{qi2020prophetnet}.



We adopt the n-stream self-attention proposed in ProphetNet \cite{qi2020prophetnet} in our decoder. The n-stream self-attention mechanism incorporates $n$ extra self-attention predicting streams besides main stream to predict next $n$ continuous future tokens respectively at each time step. 



\paragraph{Memory Scheme} ~ {To incorporate the latent variable into decoder, we adopt a memory scheme similar to OPTIMUS \cite{li2020optimus}, where latent variable $z \in \mathbb{R}^{P}$ is mapped to a additional memory vector, denoted as $h_{Mem}$, which is an additional key-value pair for decoder to attend. We have memory vector

\begin{equation}
h_{Mem}=\left[
\begin{matrix}
z_{key} \\
z_{value}
\end{matrix}
\right] = W_{M}\,z
\end{equation}

where $W_{M}\in \mathbb{R}^{H \times P}$ is the weight matrix, and the memory vector is shared and propagated across all layers in decoder as:

\begin{small}
\begin{align*}
H^{(k+1)} = \textbf{MultiHead}(H^{(k)},h_{Mem}^{(k)} \oplus H^{(k)}, h_{Mem}^{(k)} \oplus H^{(k)}) 
\end{align*}
\end{small}

where $H^{(k)}$ refers to the hidden state of the $k$-th layer of decoder. The memory vector is equivalent to adding a virtual token during decoding to participate in the calculation of self-attention main stream, and the predicting streams are implicitly affected by $h_{Mem}$ through interaction with the main stream. The latent variable guides the generation of each step of the decoder through the memory vector.







}


\subsection{Latent Variable}

Intuitively, introducing latent variables provides a hierarchical generation procedure: 1) sample a latent variable $z$ from the prior network $p(z|c)$; 2) generate $r$ through the decoder network $p(r|c,z)$. From previous research \cite{zhao-etal-2017-learning}, $z\sim p(z|c)$ may determine the high-level semantics, and the auto-regressive decoding is followed to produce the output sentences with low-level syntactic and lexical details.

Similar to the Variational Autoencoders (VAEs), we learn the parameters $\theta$ by maximizing the marginal log likelihood:
\begin{equation*}
    {\rm log}\,p_{\theta}(r|c)={\rm log}\int p_{\phi}(z|c)p_{\theta}(r|c,z)dz,
\end{equation*}

where $p_{\phi}$ involves an intractable marginalization over the latent variable $z$. 
\cite{kingma2016improving,li2020optimus}, We will optimize its lower bound, which is equivalent to minimize the two terms below: 
reconstruction loss (or negative log-likelihood)

\begin{equation}
\label{eqn:loss-rc}
\begin{aligned}
\mathcal{L}_{rc} &= - \mathbb{E}_{q(z)}[log\,p_{\theta}(r|c,z)] \\ 
   &= - \mathbb{E}_{q(z)}[log\,\prod_{t} p_{\theta}(r_{t:t+n-1}|r_{<t}, c)]
\end{aligned}
\end{equation}
 and K-L regularization term 
\begin{equation}
\label{eqn:loss-kl}
\mathcal{L}_{kl} = KL(q(z) || p_{\phi}(z|c)).
\end{equation}

Here $q(z)$ is a multivariable normal distribution with mean $\mu \in \mathbb{R}^P$ and diagonal variance matrix with diagonal taiking values $\sigma^2 \in \mathbb{R}^P$, denoted as $\text{diag}(\sigma^2)$. 

To connect to the hidden space, we add a special classification token ([CLS]) to the beginning of the context, and the first hidden state denoted as $h_{[CLS]} \in \mathbb{R}^{H}$ in last-layer is used to represent the global dialog context. We assume 

\begin{equation}
\left[
\begin{matrix}
\mu \\
log(\sigma^{2})
\end{matrix}
\right] = {{{\rm MLP}_{h}}}\,h_{[CLS]}
\end{equation}
where ${\rm MLP}_{h}$ is a multilayer perceptron and this multilayer perceptron is called the prior network in VAEs literature. We can then sample $P$ random variables with each variable is from standard normal distribution and via transformation, we obtain samples of $z\in \mathbb{R}^{P}$ from $\mathcal{N}(\mu, \text{diag}(\sigma^{2}))$, and feed them to the decoder.

\subsection{Mask Language Spans}

To improve the understanding ability of the encoder and the robustness to noise, we randomly mask part of the context before encoding. Recent research \cite{joshi2020spanbert, lewis2020bart} on masked language models show the advantages of masking spans over masking individual words or subword units. 

We adopt a simple method to mask spans: 1) randomly select $n$ tokens in context, denote as $\mathcal{S}$; 2) for each token $t \in \mathcal{S}$, extend it to a text span with a fixed length of $m$; 3) mask all selected tokens after sorting, deduplication and boundary checking.




Following BERT \cite{devlin2019bert}, the total number of masked tokens in the context accounts for approximately 15\%, and we replace the masked token with: 1) the [MASK] token 80\% of the time; 2) a random token 10\% of the time; 3) the unchanged masked token 10\% of the time. Then, the last-layer hidden states $h_{x} \in \mathbb{R}^{H}$ of each masked token $x$ will be used to predict the original token and the encoder is trained to optimize the cross entropy loss:
\begin{equation}
\label{eqn:loss-mask}
\mathcal{L}_{M} = - \sum_{x} {\rm LSM}(W_{2}\,\text{tanh}(W_{1}h_{x} + b_{1}))(x)
\end{equation}
where $W_{1} \in \mathbb{R}^{H \times H}$, $b_{1} \in \mathbb{R}^{H}$ and $W_{2} \in \mathbb{R}^{H \times |V|}$ denote the weight matrices of one fully-connected layer, $|V|$ is the vocabulary size, ${\rm LSM}$ is log softmax function and ${\rm LSM}(\dots)(x)$ means to take the log probability value corresponding to token $x$. In this paper, we share the parameters of $W_{2}$ with parameters of  embedding layers in the encoder and decoder. Note that we only mask the context only the pre-training stage. 


\subsection{Reduce KL-vanishing}

DialogVED allows the decoder to attend the hidden states of context (i.e., the output of the encoder), and thus direct training will cause the decoder to ignore the latent variable $z$, and the KL loss will rapidly decrease to 0 and the latent space loses its expressive power, which is called posterior collapse or KL-vanishing \cite{bowman2016generating}. This paper adopts two methods developed in VAEs literature to reduce posterior collapse:

\textbf{Free Bits} \cite{kingma2016improving}, which replaces the K-L regularization term in (\ref{eqn:loss-kl}) with a hinge loss term that maximize each component of the original K-L term with a constant $\lambda$:
\begin{equation}
\label{eqn:loss-kl-2}
\mathcal{L}_{kl}^{'} = -\sum_{i} max(\lambda, KL(q(z_{i}) || p_{\phi}(z_{i}|c)))
\end{equation}

\textbf{Bag-of-words Loss}~\cite{zhao2017learning}, which is used to encourage the latent variable to predict the words in response $r$ in a non-autoregressive way:
\begin{equation}
\label{eqn:loss-bow}
\mathcal{L}_{BOW} = -\sum_{t=1}^{T} log\, f_{r_{t}}
\end{equation}
where $T$ is the number of tokens in response $r$, and $f_{r_{t}}$ denotes the estimated probability of word $r_{t}$.
More specifically, $f$ is the function outputting the probability of words within the target response:
\begin{equation}
f = {\rm softmax}({{\rm MLP}_{z}}[z \oplus h_{[CLS]}]) \in \mathbb{R}^{|V|}    
\end{equation}
where ${\rm MLP}_{z}$ is a multilayer perceptron and $V$ refers to the whole vocabulary.

\subsection{Position Embeddings}
\label{section:pe}

\paragraph{Absolute Position Embeddings} \label{section:ape} ~ {Besides token-level learned position embeddings used in original Transformer, we also consider turn level and speaker-level position embeddings like PLATO \cite{bao-etal-2020-plato}. To better model the meaning of a turn in a dialog, We introduce embedding for turn position and role position in one conversation, the final input embedding of each token is the sum of corresponding turn, role and token embeddings.}



\paragraph{Relative Position Embeddings} \label{section:rpe} ~ {It has recently become more common to use relative position embeddings, which produce a different learned embedding according to the offset between the “key” and “query” being compared in the self-attention mechanism \cite{shaw2018self, raffel2019exploring}. 
We extend the element of the original relative distance matrix in T5 \cite{raffel2019exploring} to two-tuple. 
\begin{align*}
e_{ij}&=\frac{x_{i}W^{Q}(x_{j}W^{K} + a_{ij}^{K})^{T}}{\sqrt{d_{z}}}, \\
a_{ij}^{K} &= f(d_{token}, d_{turn}, x_{i}, x_{j})
\end{align*}
In the mapping function $f$, we consider both token relative distance $d_{token}$ and turn relative distance $d_{turn}$, where these tuples are mapped through a bucket function, and then $a_{ij}^{K}$ is queried in pre-defined embedding layers. 


}


\subsection{Pre-training Objectives}

Combining the losses detailed in the Equations (\ref{eqn:loss-rc}) (\ref{eqn:loss-mask}) (\ref{eqn:loss-kl-2}) and  (\ref{eqn:loss-bow}), we have pre-training objective, which we use to pre-train the DialogVED on the large-scale conversation corpus:

\begin{equation}
  loss = \mathcal{L}_{\rm{M}} + \mathcal{L}_{rc} +  \mathcal{L}_{kl}^{'} + \mathcal{L}_{BOW}  
\end{equation}
To sum up, we mask text spans in the context $c$, sample a latent variable $z$ from prior network, and then let the encoder and decoder predict the masked spans and response $r$ respectively with the guidance of the latent variable $z$. 

\section{Experiments}\label{sec.experiments}

In this section, we firstly introduce the pre-training datasets and fine-tuning benchmarks in \S~\ref{sec.datasets}, and implement details in \S~\ref{sec.model.config}. Then we present the main results in \S~\ref{sec.main.results}. Lastly, we analyze the influence of parameters and position embeddings in \S~\ref{sec.analysis}.


\subsection{DataSets and Baselines}

\label{sec.datasets}

\subsubsection{Pre-training Corpus}

Large-scale Reddit comments dataset \cite{zhou2018commonsense, galley2019grounded} is employed for pre-training our dialog language model. This dataset has been proved to be helpful in various conversation downstream tasks \cite{bao-etal-2020-plato,zhang2019dialogpt}. We use the script provided by DialoGPT \cite{zhang2019dialogpt} to obtain the latest Reddit comment data. We obtain 215 million\footnote{Given an instance containing multiple turns of dialogue $\{t_1, t_2, ..., t_n\}$, we extract $n-1$ samples (i.e. context-response pairs), where the context $c$ is $\{t_1, t_2, ..., t_{i-1}\}$, and the response $r$ is $\{t_i\}$, for $i=\{2, 3, ..., n\}$.} training samples (42GB in total) for pre-training.

\begin{table}[htbp]
\centering
\scalebox{0.65}{
\begin{tabular}{ccccr}
\toprule
Task                        & Dataset       & \# Examples   & \# Turns  & \# Tokens   \\ \midrule
\multirow{2}{*}{Pre-train}  & Reddit-Short  & 214M          & 2.6      & 28.6/16.0           \\ 
                            & Reddit-Long  & 726K          & 6.9      & 137.1/21.2          \\ \midrule\midrule
\multirow{3}{*}{Fine-tune}  & DailyDialog   & 76K           & 5.9      & 75.6/15.0           \\ 
                            & Persona-Chat  & 122K          & 8.4      & 95.1/12.2          \\ 
                            & DSTC7-AVSD    & 76K           & 10.9     & 102.1/10.7          \\ \bottomrule
\end{tabular}}
\caption{Dataset statistics used for pre-training and fine-tuning in this paper, where \# Turns means avg. turns, and \# Tokens means avg. tokens of context and response (separated by slashes) after WordPiece tokenization \cite{devlin2019bert}.}
\label{tab:datasets_stat}
\end{table}

\begin{table*}[ht]
\centering
\scalebox{0.76}{
\begin{tabular}{l|cccc|cccc}
\hline
\multicolumn{1}{c|}{\multirow{2}{*}{Model}} & \multicolumn{4}{c|}{DailyDialog}           & \multicolumn{4}{c}{PersonaChat}          \\
\multicolumn{1}{c|}{}                       & BLEU-1 & BLEU-2 & Distinct-1 & Distinct-2 & BLEU-1 & BLEU-2 &Distinct-1 & Distinct-2 \\ \hline
Seq2Seq~\cite{vinyals2015neural}            & 0.336  & 0.238  & 0.030       & 0.128      & 0.448  & 0.353  & 0.004      & 0.016      \\
iVAE\_MI~\cite{fang2019implicit}            & 0.309  & 0.249  & 0.029      & 0.250       & -      & -      & -          & -          \\
LIC~\cite{golovanov2019large}                                         & -      & -      & -          & -          & 0.405  & 0.320   & 0.019      & 0.113      \\
PLATO w/o latent  \cite{bao-etal-2020-plato}& 0.405  & 0.322  & 0.046      & 0.246      & 0.458  & 0.357  & 0.012      & 0.064      \\
PLATO \cite{bao-etal-2020-plato}            & 0.397  & 0.311  & 0.054      & 0.291      & 0.406  & 0.315  & 0.021      & 0.121      \\
ProphetNet  \cite{qi2020prophetnet}         & 0.443  & 0.392  & 0.039      & 0.211      & 0.466  & 0.391  & 0.013      & 0.075      \\ \hline
DialogVED w/o latent                        & 0.461  & 0.407  & 0.041      & 0.222      & 0.459  & 0.380  & 0.010      & 0.062      \\
DialogVED - Greedy                          & 0.459  & 0.410  & 0.045      & 0.265      & 0.470  & 0.387  & 0.016      & 0.103            \\
DialogVED - Sampling                        & 0.431  & 0.370  & \textbf{0.058} & \textbf{0.372} & 0.428 & 0.357 & \textbf{0.032} & \textbf{0.273}  \\
DialogVED                            & \textbf{0.481} & \textbf{0.421} & 0.042 & 0.232 & \textbf{0.482} & \textbf{0.399} &  0.015 & 0.094      \\ \hline      
\end{tabular}}
\caption{Experimental results on DailyDialog and PersonaChat with automatic evaluations, with highest value written in bold. The default decoding is beam search with beam = 5, and the latent size of our DialogVED is 64.}
\label{tab:dailidialog}
\end{table*}

\begin{table*}[ht]
\centering
\scalebox{0.82}{
\begin{tabular}{l|ccccccc}
\hline
Model      & BLEU-1 & BLEU-2 & BLEU-3 & BLEU-4 & METEOR & ROUGH-L & CIDEr \\ \hline
AVSD Baseline \cite{alamri2019audio}  & 0.629  & 0485   & 0.383  & 0.309  & 0.215  & 0.487  & 0.746 \\
CMU Sinbad’s  \cite{sanabria2019cmu}      & 0.718  & 0.584  & 0.478  & 0.394  & 0.267  & 0.563  & 1.094 \\
PLATO  \cite{bao-etal-2020-plato}    & 0.784  & 0.637  & 0.525  & 0.435  & 0.286  & 0.596  & 1.209 \\ 
ProphetNet \cite{qi2020prophetnet} & 0.824  & 0.691  & 0.582  & 0.487  & 0.313  & 0.635  & 1.382 \\ \hline
DialogVED w/o latent & \textbf{0.832}  & \textbf{0.705}  & \textbf{0.598} & \textbf{0.506} & \textbf{0.314} & \textbf{0.638} & 1.386        \\
DialogVED - Greedy & 0.817  & 0.685  & 0.575  & 0.481  & 0.306 & 0.629  & 1.356 \\ 
DialogVED & 0.822  & 0.692  & 0.582  & 0.489  & 0.312  &0.636  & \textbf{1.391} \\ 
 \hline
\end{tabular}}
\caption{Experimental results on DSTC7-AVSD with automatic evaluations, with highest value written in bold.}
\label{tab:dstc7-avsd}
\end{table*}

To accelerate the training process and accommodate GPU memory limitations, we adopt two methods. First, we sort the samples according to the length of the context. Samples with similar length (i.e. number of tokens in context) are assembled into a batch to minimize the amount of padding. Secondly, due to the uneven distribution of sample lengths, we divide the Reddit corpus into two sub-datasets: \textbf{Reddit-Short} and \textbf{Reddit-Long} according to the length of context and response. with some statistics in Table \ref{tab:datasets_stat}, and optimize the batch size for each sub-dataset to avoid reserving a large amount of memory for a few long response samples during the training process. Within an epoch, we first pre-train on \textbf{Reddit-Short} with a larger batch size, and then pre-train \textbf{Reddit-Long} with a smaller batch size. We split the reddit comment dataset here mainly for efficiency.

\subsubsection{Fine-tuning Benchmarks}

Following PLATO \cite{bao-etal-2020-plato}, we select three datasets as our benchmarks:

\textbf{DailyDialog}~\cite{li2017dailydialog},
a chit-chat dataset, which contains high-quality human conversations about daily life.

\textbf{Persona-Chat}~\cite{zhang2018personalizing},
a knowledge grounded conversation dataset. It provides both manually annotated conversations and corresponding persona profiles (background knowledge), where two participants chat naturally and try to get to know each other. 

\textbf{DSTC7-AVSD}~\cite{alamri2019audio}, a conversational question answering dataset, shorts for Audio Visual Scene-aware Dialog of the DSTC7 challenge. The system needs to generate an answer given dialogue context and background knowledge. There are multiple reference responses for each context in DSTC7-AVSD test set.

For evaluation, we use the same metrics as used in PLATO, except for knowledge-related metrics, since this paper does not focus on utilizing knowledge. So we will focus the following metrics:

\textbf{BLEU-1/2}~\cite{papineni2002bleu}, which measures the relevance of generated text to the reference text by calculating the 1/2-gram overlapping between them. 

\textbf{Distinct-1/2}~\cite{li2016diversity}, which measures the diversity of a generated sentence by focusing on the number of distinct 1/2-gram of a sentence and thus penalizing sentences with lots of repeated words.

Other word-overlap-based metrics, METEOR, ROUGE-L, and CIDEr, which are also reported for the DSTC7-AVSD dataset, same as DSTC7 reviews \cite{AVSD@DSTC7}.

\subsubsection{Baselines}

Vanilla sequence to sequence (Seq2Seq) models, dialog pre-training models, and general natural language pre-training models are used as our baselines: \textbf{Seq2Seq}~\cite{vinyals2015neural} is a sequence-to-sequence model with attention. $\textbf{iVAE}_{\textbf{MI}}$~\cite{fang2019implicit} is  
an implicit deep latent variable model based on Variational Autoencoder for better latent representations and diverse responses. \textbf{LIC}~\cite{golovanov2019large} 
obtains the best performance during the contest, and is one transformer based generation method.
\textbf{PLATO}~\cite{bao-etal-2020-plato}
utilizes a discrete latent variable for dialog generation pre-training to address the one-to-many problem. \textbf{ProphetNet} \cite{qi2020prophetnet} is a pre-trained LM model with predicting more than one future tokens as the pre-training objective. We fine-tune ProphetNet-Large model released in \cite{qi2020prophetnet} with downstream training data directly.

For benchmark DSTC7-AVSD, we include AVSD Baseline~\cite{alamri2019audio} system provided by the the challenge organizer, as well as the best performing model developed by the team of CMU Sinbad’s \cite{sanabria2019cmu}.

\subsection{Model Configuration}\label{sec.model.config}

DialogVED is composed of a 12-layer encoder and a 12-layer decoder, with 1024 embedding/hidden size and 4096 feed-forward filter size. The dimension $P$ of hidden states $z$ is set to 64 and we will analyze the effect of $P$ in \S~\ref{sec.sampling}. We use Adam optimizer \cite{kingma2014adam} with a learning rate of $3 \times 10^{-4} $ for pre-training. We set ngram as 2 following ProphetNet \cite{qi2020prophetnet}. The pre-training of dialogue generation is carried out on 32 Nvidia Telsa V100 32G GPU (4 nodes) for 6 epochs, taking about 5 days to reach convergence. Mixed precision training is also adopted for efficiently training and inference, and we use the Fairseq \cite{ott2019fairseq} framework to conduct all experiments. We use the BERT-uncased dictionary, and replace some unused tokens to custom special symbols (such as [SOT], denoting the beginning of the conversation, which is suitable for conversation datasets containing knowledge, like PersonaChat and DSTC7-AVSD). We used package WordPiece \cite{devlin2019bert} for tokenization.

For fine-tuning, we use exactly the same hyperparameter settings in all three datasets, and they are slightly different from the hyperparameter in pre-training. The learning rate is set to $1 \times 10^{-4}$ and the batch size is fixed to 512. We also adopt an additional warmup strategy where we linearly increase the learning rate from initial learning rate ($1 \times 10^{-7}$), the number of warmup updates is set to 2000. For each dataset, we train 10 epochs, and select the checkpoint with the lowest validation loss for inference.

\subsection{Main Results}\label{sec.main.results}


In Table~\ref{tab:dailidialog}, we compare several  DialogVED variants with baseline models. \textbf{DialogVED} represents inferencing DialogVED with beam search. Compared with DialogVED, \textbf{DialogVED w/o latent} is not equipped with latent variable, thus the loss function does not include bag-of-words loss and K-L loss. \textbf{DialogVED Greedy} means DialogVED inference with greedy search. For \textbf{DialogVED Sampling}, we sample from the top $K$ tokens with the highest output probability at each decoding step. For the latent space, we always sample each latent variable from the prior distribution standard normal distribution. Here, beam size is set to 5 and $K$ is set to 100.

As shown in Table \ref{tab:dailidialog} and Table \ref{tab:dstc7-avsd}, our model DialogVED is very competitive compared to PLATO and other models. In particular, decoding using Top-K ($K=100$) sampling with DialogVED beats the PLATO in BLEU-1/2 and Distinct-1/2 on DailyDialog and PersonaChat  (see in Table \ref{tab:dailidialog}). In fact, as $K$ increases, the overlap of n-grams decreases and the diversity increases. Based on our observations, $K$ taking 100 is a good balance, Table \ref{tab:latent_size} shows more detailed results. 

On the DSTC7-AVSD, the diversity of the responses is not as important as the accuracy. From Table \ref{tab:dstc7-avsd}, We observe that DialogVED w/o latent variable perform the best in overall metrics. However, DialogVED equipped with beam search or greedy search, can still easily beat PLATO even though it has a post-generation ranking component. 

There are 2 essential components that contribute greatly the success of our model: Firstly, We adopt a newly developed pretrained LM as the initializer and further continue its pretraining pipeline on our dialog dataset (Reddit) and thus we have a really powerful encoder-decoder. This is demonstrated in the fact that our model (DialogVED w/o latent variable) beat PLATO (w/o latent variable) in all metrics on all the three datasets.

Secondly, the special structure of our model combines the benefits of both seq2seq models and VAE models. Compared to general VAEs, DialogVED allows encoder-decoder interaction in the decoding, which avoids insufficient representation of low-dimensional latent variable. At the same time, compared with seq2seq model, predicting the bag of words pushes the latent variable to give extra guidance to decoder. This is demonstrated by the fact that when compared with DialogVED w/o latent variable, we observe the additional gains in terms of both accuracy and diversity (see Table \ref{tab:dailidialog}). 

Overall, our DialogVED achieves new state-of-the-art results in all three downstream tasks of dialogue response generation. 


\subsection{Parameters and Position Analysis}
\label{sec.analysis}
\subsubsection{Balancing Accuracy and Diversity with Sampling} \label{sec.sampling}
We investigate the effect of latent space sizes, $P$, defined as the dimension of the latent variable $z$ and the different $K$ in sampling. 

The results in Table~\ref{tab:latent_size} show that smaller latent size ($P=32$) is more dominant in n-gram based metrics (BLEU-1/2), while larger latent size generates more diverse texts. From the results of top-$K$ sampling, we see that the two metric (BLEU-1/2 and Distinct-1/2) have a negative correlation.

We can flexibly choose the decoding strategy depends on specific scene.

\begin{table}[htbp]
\centering
\scalebox{0.8}{
\begin{tabular}{c|c|c|c|c|c}
\hline
$P$                     & Top-$K$  & BLEU-1& BLEU-2      & Distinct-1& Distinct-2 \\ \hline
\multirow{4}{*}{32}     & 5        & \textbf{0.448} & \textbf{0.385}       & 0.042 & 0.289           \\ 
                        & 20       & 0.443 & 0.376       & 0.045 & 0.317           \\ 
                        & 50       & 0.442 & 0.375       & 0.047 & 0.332           \\ 
                        & 100      & 0.439 & 0.374       & 0.051 & 0.347           \\ \hline\hline
\multirow{4}{*}{64}     & 5        & 0.442 & 0.383       & 0.046 & 0.308           \\ 
                        & 20       & 0.437 & 0.374       & 0.050    &0.340           \\ 
                        & 50       & 0.434 & 0.371       & 0.054&0.364           \\ 
                        & 100      & 0.431 & 0.370       & \textbf{0.058} & \textbf{0.372}          \\ \hline
\end{tabular}}
\caption{The results of different latent size and Top $K$ on DailyDialog dataset. $P$ is the dimension of the latent variable.}
\label{tab:latent_size}
\end{table}

\subsubsection{Position Embeddings}\label{sec.positions}

We study the impact of position embeddings as described in section~\ref{section:pe}, we define two types of position embeddings: absolute position embeddings (APE) and relative position embeddings (RPE). We report the metrics of their different combinations, these independent components are TurnAPE (turn absolute embedding), RoleAPE (role absolute embedding), TokenRPE (token relative embedding) and TurnRPE(turn relative embedding) respectively. 




As the results shown in Table~\ref{tab:ape_rpe}, the combination of TurnAPE and RoleAPE achieve the best performance. Both absolute and relative position embeddings improve model performance, nevertheless, including them at the same time can be harmful. 

\begin{table}[h]
\centering
\scalebox{0.85}{
\begin{tabular}{c|c|c|c|c|c}
\hline
\begin{tabular}[c]{@{}l@{}}Turn\\ APE\end{tabular} & \begin{tabular}[c]{@{}l@{}}Role\\ APE\end{tabular} & \begin{tabular}[c]{@{}l@{}}Token\\ RPE\end{tabular} & \begin{tabular}[c]{@{}l@{}}Turn\\ RPE\end{tabular} & BLEU-1/2 & Distinct-1/2 \\ \hline
         &          &          &          & 0.481/0.421 & \textbf{0.042/0.232}  \\ \hline
\checkmark       &          &          &          & 0.491/0.429 & 0.041/0.229  \\ \hline
         & \checkmark       &          &          & 0.483/0.422 & 0.042/0.230  \\ \hline
\checkmark       & \checkmark       &          &          & \textbf{0.494/0.435} & 0.042/0.232  \\ \hline
         &          & \checkmark       &          & 0.485/0.424 & 0.039/0.216  \\ \hline
         &          &          & \checkmark       & 0.480/0.422 & 0.042/0.231  \\ \hline
         &          & \checkmark       & \checkmark       & 0.487/0.425 & 0.040/0.230  \\ \hline
\checkmark       & \checkmark       & \checkmark       & \checkmark       & 0.483/0.435 & 0.039/0.228  \\ \hline
\end{tabular}}
\caption{Effect of position embeddings on model performance.}
\label{tab:ape_rpe}
\end{table}

\begin{table}[]
\centering
\scalebox{0.9}{
\begin{tabular}{lcc|cc} \hline
\multirow{2}{*}{}        & \multicolumn{2}{c|}{Group 1}        & \multicolumn{2}{c}{Group 2}        \\
                         & Win & Lose & Win & Lose \\ \hline
Fluency         & \textbf{0.16}  & 0.11  & \textbf{0.19}  & 0.13            \\
Coherence        & \textbf{0.38} & 0.16  & 0.22  & \textbf{0.24}            \\
Informativeness & \textbf{0.13}  & 0.11 & \textbf{0.31}  & 0.14             \\
Overall        & \textbf{0.26}  & 0.14 & \textbf{0.24}  & 0.17            \\ \hline
\end{tabular}}
\caption{Human evaluation of DialogVED vs. PLATO (Group 1) and DialogVED-sampling vs. PLATO (Group 2).}
\label{tab:compare}
\end{table}

\subsection{Human Evaluation}

Automated metrics (BLEU 1/2, Distinct-1/2, etc.) have limitations for evaluating open-domain dialog tasks. To make it more convincing, we conduct a human evaluation. Specifically, we randomly select 100 dialogue contexts and generate responses with the following methods: PLATO, DialogVED and DialogVED-Sampling. Following PLATO, annotators are asked to compare the response (win, tie or lose) quality from four aspects: fluency, coherence, informativeness and overall.

The results of human comparison are shown in Table \ref{tab:compare}, where the average Cohen's kappa \cite{kraemer2014kappa} of group 1 and 2 is 0.729 and 0.743 respectively, indicating annotators have reached moderate agreement. It can be seen that most of the time they are tied, and the three models sometimes generate exactly the same response. For DialogVED, it beats Plato more in coherence but with close informativeness; while DialogVED-sampling beats Plato significantly in informativeness but with a slightly weaker coherence. 


In general, DialogVED can generate both relevant and diverse response, we show some case study to help illustrate the effectiveness of our model in Appendix \ref{sec:case_study}.
 
 

\section{Related Work}
\label{section:related}

\paragraph{Encoder-Decoder dialog models} ~ {Unlike retrieval based dialogue systems \cite{boussaha2019deep,chen2021contextual}, encoder-decoder models are widely used in dialog response generation, but it tends to generate generic responses and dull responses (e.g., I don’t know). To enhance encoder-decode models and generate diverse responses, researchers have tried different approaches: using diversity promotion objectives \cite{li2016diversity}, using different decoding algorithms \cite{li2016simple}, adding additional contents \cite{xu-etal-2019-neural}, or introducing large-scale knowledge graphs into dialog generation  \cite{liu2018knowledge, wu2020diverse}.

Another class of methods is using the latent variable to address the one-to-many problem in response generation. These models introduce discourse-level diversity and are able to generate diverse dialog responses \cite{serban2017hierarchical, zhao-etal-2017-learning, zhao-etal-2018-unsupervised, gao2019generating}. In this paper, we also adopt this approach and further we incorporate the latent variables both in the pre-training and fine-tuning.

}

\paragraph{Pre-trained Dialog Models} ~ {
Pre-trained language models have been successfully used in NLG and NLU tasks \cite{devlin2019bert,radford2019language}.  
Recently, various new pre-trained language models have been pre-trained including BART \cite{lewis2020bart}, ProphetNet \cite{qi2020prophetnet}, T5 \cite{raffel2020exploring}. In these papers, they demonstrate that better performance can be obtained with fine-tuning PLMs than training from scratch.

Due to the fact that there are many important applications in the dialog domain and the dialog corpus has different linguistic features from general documents, pre-trained dialog models with open domain dialog data such as Reddit is very important. DialoGPT \cite{zhang2019dialogpt} continues to pre-train GPT-2 model directly on Reddit comments data, and the new pre-trained model achieves better performance on downstream tasks including several dialog response generation benchmarks.  

PLATO \cite{bao-etal-2020-plato} proposes a new model specifically for dialog generation, which introduces a discrete variable for one-to-many relationship modeling. The pre-trained model helps to achieve state-of-the-art results on several response generation tasks. This is the closest work in literature to ours. However, in our paper, we introduce continuous latent variables during pre-training on dialog corpus instead of a discrete latent variable. 
}

\section{Conclusion}
\label{section:conclusions}

This paper proposes a new pre-training framework for dialogue response generation called DialogVED. The latent variable is incorporated into the sequence-to-sequence framework based on Transformer, and obtains a robust and diverse response generation model through 4 training targets. our pre-trained model has achieved new state-of-the-art in multiple downstream tasks of dialogue response generation. Extensive experiments prove the effectiveness of our model. Additional human evaluation demonstrates the advantages of our proposed model. 



\section*{Ethical Statement}

In this paper, different ethical restrictions deserve discussion.

All data used in our pre-training are available online and other dialog corpus in this paper are publicly available sources. We strictly followed the platform's policies and rules when crawling data from web platforms. We did not employ any author-specific information in our research.

Our corpus may includes some bias, such as political bias and social bias, and our model might have inherited some forms of these bias. In order to limit these bias as much as possible, we filter controversial articles and removed data with offensive information when possible.

\bibliography{anthology,custom}

\begin{thebibliography}{46}
\expandafter\ifx\csname natexlab\endcsname\relax\def\natexlab#1{#1}\fi

\bibitem[{Alamri et~al.(2019{\natexlab{a}})Alamri, Cartillier, Das, Wang,
  Cherian, Essa, Batra, Marks, Hori, Anderson et~al.}]{alamri2019audio}
Huda Alamri, Vincent Cartillier, Abhishek Das, Jue Wang, Anoop Cherian, Irfan
  Essa, Dhruv Batra, Tim~K Marks, Chiori Hori, Peter Anderson, et~al.
  2019{\natexlab{a}}.
\newblock Audio visual scene-aware dialog.
\newblock In \emph{Proceedings of the IEEE/CVF Conference on Computer Vision
  and Pattern Recognition}, pages 7558--7567.

\bibitem[{Alamri et~al.(2019{\natexlab{b}})Alamri, Hori, Marks, Batra, and
  Parikh}]{AVSD@DSTC7}
Huda Alamri, Chiori Hori, Tim~K Marks, Dhruv Batra, and Devi Parikh.
  2019{\natexlab{b}}.
\newblock Audio visual scene-aware dialog ({AVSD}) track for natural language
  generation in {DSTC7}.
\newblock In \emph{AAAI workshop on the 7th edition of Dialog System Technology
  Challenge (DSTC7)}.

\bibitem[{Bahuleyan et~al.(2017)Bahuleyan, Mou, Vechtomova, and
  Poupart}]{bahuleyan2017variational}
Hareesh Bahuleyan, Lili Mou, Olga Vechtomova, and Pascal Poupart. 2017.
\newblock Variational attention for sequence-to-sequence models.
\newblock \emph{arXiv preprint arXiv:1712.08207}.

\bibitem[{Bao et~al.(2020)Bao, He, Wang, Wu, and Wang}]{bao-etal-2020-plato}
Siqi Bao, Huang He, Fan Wang, Hua Wu, and Haifeng Wang. 2020.
\newblock \href {https://doi.org/10.18653/v1/2020.acl-main.9} {{PLATO}:
  Pre-trained dialogue generation model with discrete latent variable}.
\newblock In \emph{Proceedings of the 58th Annual Meeting of the Association
  for Computational Linguistics}, pages 85--96, Online. Association for
  Computational Linguistics.

\bibitem[{Boussaha et~al.(2019)Boussaha, Hernandez, Jacquin, and
  Morin}]{boussaha2019deep}
Basma El~Amel Boussaha, Nicolas Hernandez, Christine Jacquin, and Emmanuel
  Morin. 2019.
\newblock Deep retrieval-based dialogue systems: A short review.
\newblock \emph{arXiv preprint arXiv:1907.12878}.

\bibitem[{Bowman et~al.(2016)Bowman, Vilnis, Vinyals, Dai, Jozefowicz, and
  Bengio}]{bowman2016generating}
Samuel Bowman, Luke Vilnis, Oriol Vinyals, Andrew Dai, Rafal Jozefowicz, and
  Samy Bengio. 2016.
\newblock Generating sentences from a continuous space.
\newblock In \emph{Proceedings of The 20th SIGNLL Conference on Computational
  Natural Language Learning}, pages 10--21.

\bibitem[{Chen et~al.(2021)Chen, Gong, Xu, Hu, Yao, Wei, Fan, Hu, Zhou, Cheng
  et~al.}]{chen2021contextual}
Wei Chen, Yeyun Gong, Can Xu, Huang Hu, Bolun Yao, Zhongyu Wei, Zhihao Fan,
  Xiaowu Hu, Bartuer Zhou, Biao Cheng, et~al. 2021.
\newblock Contextual fine-to-coarse distillation for coarse-grained response
  selection in open-domain conversations.
\newblock \emph{arXiv preprint arXiv:2109.13087}.

\bibitem[{Devlin et~al.(2019)Devlin, Chang, Lee, and
  Toutanova}]{devlin2019bert}
Jacob Devlin, Ming-Wei Chang, Kenton Lee, and Kristina Toutanova. 2019.
\newblock Bert: Pre-training of deep bidirectional transformers for language
  understanding.
\newblock In \emph{Proceedings of the 2019 Conference of the North American
  Chapter of the Association for Computational Linguistics: Human Language
  Technologies, Volume 1 (Long and Short Papers)}, pages 4171--4186.

\bibitem[{Fang et~al.(2019)Fang, Li, Gao, Dong, and Chen}]{fang2019implicit}
Le~Fang, Chunyuan Li, Jianfeng Gao, Wen Dong, and Changyou Chen. 2019.
\newblock Implicit deep latent variable models for text generation.
\newblock In \emph{Proceedings of the 2019 Conference on Empirical Methods in
  Natural Language Processing and the 9th International Joint Conference on
  Natural Language Processing (EMNLP-IJCNLP)}, pages 3937--3947.

\bibitem[{Galley et~al.(2019)Galley, Brockett, Gao, Gao, and
  Dolan}]{galley2019grounded}
Michel Galley, Chris Brockett, Xiang Gao, Jianfeng Gao, and Bill Dolan. 2019.
\newblock Grounded response generation task at dstc7.
\newblock In \emph{AAAI Dialog System Technology Challenges Workshop}.

\bibitem[{Gao et~al.(2019)Gao, Bi, Liu, Li, and Shi}]{gao2019generating}
Jun Gao, Wei Bi, Xiaojiang Liu, Junhui Li, and Shuming Shi. 2019.
\newblock Generating multiple diverse responses for short-text conversation.
\newblock In \emph{Proceedings of the AAAI Conference on Artificial
  Intelligence}, volume~33, pages 6383--6390.

\bibitem[{Golovanov et~al.(2019)Golovanov, Kurbanov, Nikolenko, Truskovskyi,
  Tselousov, and Wolf}]{golovanov2019large}
Sergey Golovanov, Rauf Kurbanov, Sergey Nikolenko, Kyryl Truskovskyi, Alexander
  Tselousov, and Thomas Wolf. 2019.
\newblock Large-scale transfer learning for natural language generation.
\newblock In \emph{Proceedings of the 57th Annual Meeting of the Association
  for Computational Linguistics}, pages 6053--6058.

\bibitem[{Joshi et~al.(2020)Joshi, Chen, Liu, Weld, Zettlemoyer, and
  Levy}]{joshi2020spanbert}
Mandar Joshi, Danqi Chen, Yinhan Liu, Daniel~S Weld, Luke Zettlemoyer, and Omer
  Levy. 2020.
\newblock Spanbert: Improving pre-training by representing and predicting
  spans.
\newblock \emph{Transactions of the Association for Computational Linguistics},
  8:64--77.

\bibitem[{Kingma and Ba(2014)}]{kingma2014adam}
Diederik~P Kingma and Jimmy Ba. 2014.
\newblock Adam: A method for stochastic optimization.
\newblock \emph{arXiv preprint arXiv:1412.6980}.

\bibitem[{Kingma et~al.(2016)Kingma, Salimans, Jozefowicz, Chen, Sutskever, and
  Welling}]{kingma2016improving}
Diederik~P Kingma, Tim Salimans, Rafal Jozefowicz, Xi~Chen, Ilya Sutskever, and
  Max Welling. 2016.
\newblock Improving variational inference with inverse autoregressive flow.
\newblock \emph{arXiv preprint arXiv:1606.04934}.

\bibitem[{Kraemer(2014)}]{kraemer2014kappa}
Helena~C Kraemer. 2014.
\newblock Kappa coefficient.
\newblock \emph{Wiley StatsRef: Statistics Reference Online}, pages 1--4.

\bibitem[{Lewis et~al.(2020)Lewis, Liu, Goyal, Ghazvininejad, Mohamed, Levy,
  Stoyanov, and Zettlemoyer}]{lewis2020bart}
Mike Lewis, Yinhan Liu, Naman Goyal, Marjan Ghazvininejad, Abdelrahman Mohamed,
  Omer Levy, Veselin Stoyanov, and Luke Zettlemoyer. 2020.
\newblock Bart: Denoising sequence-to-sequence pre-training for natural
  language generation, translation, and comprehension.
\newblock In \emph{Proceedings of the 58th Annual Meeting of the Association
  for Computational Linguistics}, pages 7871--7880.

\bibitem[{Li et~al.(2020)Li, Gao, Li, Li, Peng, Zhang, and Gao}]{li2020optimus}
Chunyuan Li, Xiang Gao, Yuan Li, Xiujun Li, Baolin Peng, Yizhe Zhang, and
  Jianfeng Gao. 2020.
\newblock Optimus: Organizing sentences via pre-trained modeling of a latent
  space.
\newblock \emph{arXiv preprint arXiv:2004.04092}.

\bibitem[{Li et~al.(2016{\natexlab{a}})Li, Galley, Brockett, Gao, and
  Dolan}]{li2016diversity}
Jiwei Li, Michel Galley, Chris Brockett, Jianfeng Gao, and William~B Dolan.
  2016{\natexlab{a}}.
\newblock A diversity-promoting objective function for neural conversation
  models.
\newblock In \emph{Proceedings of the 2016 Conference of the North American
  Chapter of the Association for Computational Linguistics: Human Language
  Technologies}, pages 110--119.

\bibitem[{Li et~al.(2016{\natexlab{b}})Li, Monroe, and Jurafsky}]{li2016simple}
Jiwei Li, Will Monroe, and Dan Jurafsky. 2016{\natexlab{b}}.
\newblock A simple, fast diverse decoding algorithm for neural generation.
\newblock \emph{arXiv preprint arXiv:1611.08562}.

\bibitem[{Li et~al.(2017)Li, Su, Shen, Li, Cao, and Niu}]{li2017dailydialog}
Yanran Li, Hui Su, Xiaoyu Shen, Wenjie Li, Ziqiang Cao, and Shuzi Niu. 2017.
\newblock Dailydialog: A manually labelled multi-turn dialogue dataset.
\newblock In \emph{Proceedings of the Eighth International Joint Conference on
  Natural Language Processing (Volume 1: Long Papers)}, pages 986--995.

\bibitem[{Lin et~al.(2020)Lin, Madotto, and Fung}]{lin2020exploring}
Zhaojiang Lin, Andrea Madotto, and Pascale Fung. 2020.
\newblock Exploring versatile generative language model via parameter-efficient
  transfer learning.
\newblock In \emph{Proceedings of the 2020 Conference on Empirical Methods in
  Natural Language Processing: Findings}, pages 441--459.

\bibitem[{Liu et~al.(2018)Liu, Chen, Ren, Feng, Liu, and
  Yin}]{liu2018knowledge}
Shuman Liu, Hongshen Chen, Zhaochun Ren, Yang Feng, Qun Liu, and Dawei Yin.
  2018.
\newblock Knowledge diffusion for neural dialogue generation.
\newblock In \emph{Proceedings of the 56th Annual Meeting of the Association
  for Computational Linguistics (Volume 1: Long Papers)}, pages 1489--1498.

\bibitem[{Luong et~al.(2015)Luong, Pham, and Manning}]{luong2015effective}
Minh-Thang Luong, Hieu Pham, and Christopher~D Manning. 2015.
\newblock Effective approaches to attention-based neural machine translation.
\newblock \emph{arXiv preprint arXiv:1508.04025}.

\bibitem[{Oord et~al.(2017)Oord, Vinyals, and Kavukcuoglu}]{oord2017neural}
Aaron van~den Oord, Oriol Vinyals, and Koray Kavukcuoglu. 2017.
\newblock Neural discrete representation learning.
\newblock \emph{arXiv preprint arXiv:1711.00937}.

\bibitem[{Ott et~al.(2019)Ott, Edunov, Baevski, Fan, Gross, Ng, Grangier, and
  Auli}]{ott2019fairseq}
Myle Ott, Sergey Edunov, Alexei Baevski, Angela Fan, Sam Gross, Nathan Ng,
  David Grangier, and Michael Auli. 2019.
\newblock fairseq: A fast, extensible toolkit for sequence modeling.
\newblock \emph{arXiv preprint arXiv:1904.01038}.

\bibitem[{Papineni et~al.(2002)Papineni, Roukos, Ward, and
  Zhu}]{papineni2002bleu}
Kishore Papineni, Salim Roukos, Todd Ward, and Wei-Jing Zhu. 2002.
\newblock Bleu: a method for automatic evaluation of machine translation.
\newblock In \emph{Proceedings of the 40th annual meeting of the Association
  for Computational Linguistics}, pages 311--318.

\bibitem[{Qi et~al.(2020)Qi, Yan, Gong, Liu, Duan, Chen, Zhang, and
  Zhou}]{qi2020prophetnet}
Weizhen Qi, Yu~Yan, Yeyun Gong, Dayiheng Liu, Nan Duan, Jiusheng Chen, Ruofei
  Zhang, and Ming Zhou. 2020.
\newblock Prophetnet: Predicting future n-gram for sequence-to-sequence
  pre-training.
\newblock In \emph{Proceedings of the 2020 Conference on Empirical Methods in
  Natural Language Processing: Findings}, pages 2401--2410.

\bibitem[{Radford et~al.(2019)Radford, Wu, Child, Luan, Amodei, and
  Sutskever}]{radford2019language}
Alec Radford, Jeff Wu, Rewon Child, David Luan, Dario Amodei, and Ilya
  Sutskever. 2019.
\newblock Language models are unsupervised multitask learners.

\bibitem[{Raffel et~al.(2019)Raffel, Shazeer, Roberts, Lee, Narang, Matena,
  Zhou, Li, and Liu}]{raffel2019exploring}
Colin Raffel, Noam Shazeer, Adam Roberts, Katherine Lee, Sharan Narang, Michael
  Matena, Yanqi Zhou, Wei Li, and Peter~J Liu. 2019.
\newblock Exploring the limits of transfer learning with a unified text-to-text
  transformer.
\newblock \emph{arXiv preprint arXiv:1910.10683}.

\bibitem[{Raffel et~al.(2020)Raffel, Shazeer, Roberts, Lee, Narang, Matena,
  Zhou, Li, and Liu}]{raffel2020exploring}
Colin Raffel, Noam Shazeer, Adam Roberts, Katherine Lee, Sharan Narang, Michael
  Matena, Yanqi Zhou, Wei Li, and Peter~J Liu. 2020.
\newblock Exploring the limits of transfer learning with a unified text-to-text
  transformer.
\newblock \emph{Journal of Machine Learning Research}, 21:1--67.

\bibitem[{Sanabria et~al.(2019)Sanabria, Palaskar, and Metze}]{sanabria2019cmu}
Ramon Sanabria, Shruti Palaskar, and Florian Metze. 2019.
\newblock Cmu sinbad’s submission for the dstc7 avsd challenge.

\bibitem[{Serban et~al.(2017)Serban, Sordoni, Lowe, Charlin, Pineau, Courville,
  and Bengio}]{serban2017hierarchical}
Iulian Serban, Alessandro Sordoni, Ryan Lowe, Laurent Charlin, Joelle Pineau,
  Aaron Courville, and Yoshua Bengio. 2017.
\newblock A hierarchical latent variable encoder-decoder model for generating
  dialogues.
\newblock In \emph{Proceedings of the AAAI Conference on Artificial
  Intelligence}, volume~31.

\bibitem[{Shaw et~al.(2018)Shaw, Uszkoreit, and Vaswani}]{shaw2018self}
Peter Shaw, Jakob Uszkoreit, and Ashish Vaswani. 2018.
\newblock Self-attention with relative position representations.
\newblock \emph{arXiv preprint arXiv:1803.02155}.

\bibitem[{Vahdat et~al.(2018)Vahdat, Macready, Bian, Khoshaman, and
  Andriyash}]{vahdat2018dvae++}
Arash Vahdat, William Macready, Zhengbing Bian, Amir Khoshaman, and Evgeny
  Andriyash. 2018.
\newblock Dvae++: Discrete variational autoencoders with overlapping
  transformations.
\newblock In \emph{International Conference on Machine Learning}, pages
  5035--5044. PMLR.

\bibitem[{Vaswani et~al.(2017)Vaswani, Shazeer, Parmar, Uszkoreit, Jones,
  Gomez, Kaiser, and Polosukhin}]{vaswani2017attention}
Ashish Vaswani, Noam Shazeer, Niki Parmar, Jakob Uszkoreit, Llion Jones,
  Aidan~N Gomez, Lukasz Kaiser, and Illia Polosukhin. 2017.
\newblock Attention is all you need.
\newblock \emph{arXiv preprint arXiv:1706.03762}.

\bibitem[{Vinyals and Le(2015)}]{vinyals2015neural}
Oriol Vinyals and Quoc Le. 2015.
\newblock A neural conversational model.
\newblock \emph{arXiv preprint arXiv:1506.05869}.

\bibitem[{Wu et~al.(2020)Wu, Li, Zhang, Zhou, and Wu}]{wu2020diverse}
Sixing Wu, Ying Li, Dawei Zhang, Yang Zhou, and Zhonghai Wu. 2020.
\newblock Diverse and informative dialogue generation with context-specific
  commonsense knowledge awareness.
\newblock In \emph{Proceedings of the 58th annual meeting of the association
  for computational linguistics}, pages 5811--5820.

\bibitem[{Xiao et~al.(2020)Xiao, Zhang, Li, Sun, Tian, Wu, and
  Wang}]{xiao2020ernie}
Dongling Xiao, Han Zhang, Yukun Li, Yu~Sun, Hao Tian, Hua Wu, and Haifeng Wang.
  2020.
\newblock Ernie-gen: An enhanced multi-flow pre-training and fine-tuning
  framework for natural language generation.
\newblock \emph{arXiv preprint arXiv:2001.11314}.

\bibitem[{Xu et~al.(2019)Xu, Wu, Tao, Hu, Schuerman, and
  Wang}]{xu-etal-2019-neural}
Can Xu, Wei Wu, Chongyang Tao, Huang Hu, Matt Schuerman, and Ying Wang. 2019.
\newblock \href {https://doi.org/10.18653/v1/P19-1538} {Neural response
  generation with meta-words}.
\newblock In \emph{Proceedings of the 57th Annual Meeting of the Association
  for Computational Linguistics}, pages 5416--5426, Florence, Italy.
  Association for Computational Linguistics.

\bibitem[{Zhang et~al.(2018)Zhang, Dinan, Urbanek, Szlam, Kiela, and
  Weston}]{zhang2018personalizing}
Saizheng Zhang, Emily Dinan, Jack Urbanek, Arthur Szlam, Douwe Kiela, and Jason
  Weston. 2018.
\newblock Personalizing dialogue agents: I have a dog, do you have pets too?
\newblock In \emph{ACL (1)}.

\bibitem[{Zhang et~al.(2020)Zhang, Sun, Galley, Chen, Brockett, Gao, Gao, Liu,
  and Dolan}]{zhang2019dialogpt}
Yizhe Zhang, Siqi Sun, Michel Galley, Yen-Chun Chen, Chris Brockett, Xiang Gao,
  Jianfeng Gao, Jingjing Liu, and Bill Dolan. 2020.
\newblock Dialogpt: Large-scale generative pre-training for conversational
  response generation.
\newblock In \emph{ACL, system demonstration}.

\bibitem[{Zhao et~al.(2018)Zhao, Lee, and
  Eskenazi}]{zhao-etal-2018-unsupervised}
Tiancheng Zhao, Kyusong Lee, and Maxine Eskenazi. 2018.
\newblock \href {https://doi.org/10.18653/v1/P18-1101} {Unsupervised discrete
  sentence representation learning for interpretable neural dialog generation}.
\newblock In \emph{Proceedings of the 56th Annual Meeting of the Association
  for Computational Linguistics (Volume 1: Long Papers)}, pages 1098--1107,
  Melbourne, Australia. Association for Computational Linguistics.

\bibitem[{Zhao et~al.(2017{\natexlab{a}})Zhao, Zhao, and
  Eskenazi}]{zhao-etal-2017-learning}
Tiancheng Zhao, Ran Zhao, and Maxine Eskenazi. 2017{\natexlab{a}}.
\newblock \href {https://doi.org/10.18653/v1/P17-1061} {Learning
  discourse-level diversity for neural dialog models using conditional
  variational autoencoders}.
\newblock In \emph{Proceedings of the 55th Annual Meeting of the Association
  for Computational Linguistics (Volume 1: Long Papers)}, pages 654--664,
  Vancouver, Canada. Association for Computational Linguistics.

\bibitem[{Zhao et~al.(2017{\natexlab{b}})Zhao, Zhao, and
  Eskenazi}]{zhao2017learning}
Tiancheng Zhao, Ran Zhao, and Maxine Eskenazi. 2017{\natexlab{b}}.
\newblock Learning discourse-level diversity for neural dialog models using
  conditional variational autoencoders.
\newblock In \emph{Proceedings of the 55th Annual Meeting of the Association
  for Computational Linguistics (Volume 1: Long Papers)}, pages 654--664.

\bibitem[{Zhou et~al.(2018)Zhou, Young, Huang, Zhao, Xu, and
  Zhu}]{zhou2018commonsense}
Hao Zhou, Tom Young, Minlie Huang, Haizhou Zhao, Jingfang Xu, and Xiaoyan Zhu.
  2018.
\newblock Commonsense knowledge aware conversation generation with graph
  attention.
\newblock In \emph{IJCAI}, pages 4623--4629.

\end{thebibliography}
\bibliographystyle{acl_natbib}

\appendix

\section{Case Study}
\label{sec:case_study}

We demonstrate the responses generated from our model as well as other baseline models in Table~\ref{tab:case_study_dailydialog}, ~\ref{tab:case_study_personachat} and~\ref{tab:case_study_dstc7avsd}, respectively. The results in Table~\ref{tab:case_study_personachat} and ~\ref{tab:case_study_dstc7avsd} show that our model accurately outputs the knowledge information contained in context although we do not model knowledge explicitly. Compared with beam search or greedy decoding, decoding with top-K sampling not only generates bolder and more diverse response, but also can maintain good relevance, as showed in Table~\ref{tab:case_study_dailydialog} and ~\ref{tab:case_study_personachat}.

\begin{table*}
\centering
\scalebox{0.9}{
\begin{tabular}{c|c}
\hline
Context
& \multicolumn{1}{|p{12cm}}{
[P2] have you seen the new cosmo magazine that i was reading this morning ?

[P1] i think i may have seen it on the kitchen table .

[P2] you're right , thanks . there's an interesting interview with a top director about his latest film that i want to finish .

[P1] who's the director ?
} 
\\ \hline
GOLD
& \multicolumn{1}{|p{12cm}}{
it ' s martin scorsese . you liked his film , the departed , didn ' t you ?
}
\\ \hline \hline
\textbf{Model} & \textbf{Response} \\ \hline
Seq2Seq
& \multicolumn{1}{|p{12cm}}{
yes , i did . i found it very disappointing .
} 
\\ \hline
${\rm iVAE_{MI}}$
& \multicolumn{1}{|p{12cm}}{
the cake and is very popular .
}
\\ \hline
PLATO
& \multicolumn{1}{|p{12cm}}{
i'm not sure , but i think he might be the director of gone with the wind   .
}
\\ \hline \hline
DialogVED
& \multicolumn{1}{|p{12cm}}{
it ' s the director of a foreign film called ' gone with the wind ' .
}
\\ \hline
DialogVED-Greedy
& \multicolumn{1}{|p{12cm}}{
it ' s directed by a korean director , named zhuang lingo .
}
\\ \hline
DialogVED-Sampling
& \multicolumn{1}{|p{12cm}}{
steven spielberg . he said that the film will have a similar target audience as the titanic .
}
\\ \hline
\end{tabular}
}
\caption{Case analysis of response generation on DailyDialog}
\label{tab:case_study_dailydialog}
\end{table*}

\begin{table*}[htbp]
\centering
\scalebox{0.90}{
\begin{tabular}{c|c}
\hline
Persona Profiles
& \multicolumn{1}{|p{12cm}}{
i love trying barbeque restaurants . 

i am an administrative assistant . 

i am looking forward to retiring in five years .

i like ot play racquetball .
} 
\\ \hline
Context
& \multicolumn{1}{|p{12cm}}{
[P2] hi , do you like bbq ?

[P1] hello yes i love bbq

[P2] i love restaurants with bbq , they are a lot of fun .

[P1] my poor toes are so webbed

[P2] really ? afterwards , i go and play racquetball to loose some calories .

[P1] cool i like to paint

[P2] what do you do as a profession ? i work as administrative assistant

[P1] i dig in people trash for things to build

[P2] sounds unique but that is more of a hobby instead of a profession

[P1] true . i paint for a living
} 
\\ \hline
GOLD
& \multicolumn{1}{|p{12cm}}{
good thing i am going to retire in 5 years , no more working for this guy !
}
\\ \hline \hline
\textbf{Model} & \textbf{Response} \\ \hline
Seq2Seq
& \multicolumn{1}{|p{12cm}}{
that sounds like a lot of work . do you have a job ?
} 
\\ \hline
LIC
& \multicolumn{1}{|p{12cm}}{
yeah it is a lot of fun
}
\\ \hline
PLATO
& \multicolumn{1}{|p{12cm}}{
that is cool , i am looking forward to retiring in 5 years
}
\\ \hline \hline
DialogVED
& \multicolumn{1}{|p{12cm}}{
i am looking forward to retiring in five years .
}
\\ \hline
DialogVED-Greedy
& \multicolumn{1}{|p{12cm}}{
i am looking forward to retiring in five years .
}
\\ \hline
DialogVED-Sampling
& \multicolumn{1}{|p{12cm}}{
hmmm . is that what you do after you retire ? i am looking forward to retiring in five years .
}
\\ \hline
\end{tabular}
}
\caption{Case analysis of response generation on PersonaChat}
\label{tab:case_study_personachat}
\end{table*}

\begin{table*}
\centering
\scalebox{0.90}{
\begin{tabular}{c|c}
\hline
Video Caption
& \multicolumn{1}{|p{12cm}}{
a man closes his window , then he sneezes twice before taking a drink . then he opens up a bag and digs through it looking for something before walking out of the room .
} 
\\ \hline
Video Summary
& \multicolumn{1}{|p{12cm}}{
a man closes the window , goes to the table and goes through the items in a bag , takes a drink from the green cup and leaves the room .
} 
\\ \hline
Context
& \multicolumn{1}{|p{12cm}}{
[P1] what is the guy doing at the window ?

[P2] the guy is closing the window

[P1] what does he do after that ?
} 
\\ \hline
GOLD
& \multicolumn{1}{|p{12cm}}{
stands at the table and takes things out of bag
}
\\ \hline \hline
\textbf{Model} & \textbf{Response} \\ \hline
Baseline
& \multicolumn{1}{|p{12cm}}{
he picks up a book from the table
} 
\\ \hline
PLATO
& \multicolumn{1}{|p{12cm}}{
he goes to the table and takes a drink from a green cup
}
\\ \hline \hline
DialogVED
& \multicolumn{1}{|p{12cm}}{
he goes to the table and goes through the items in a bag before taking a drink
}
\\ \hline
DialogVED-Greedy
& \multicolumn{1}{|p{12cm}}{
he sneezes twice before taking a drink
}
\\ \hline
\end{tabular}
}
\caption{Case analysis of response generation on DSTC7-AVSD}
\label{tab:case_study_dstc7avsd}
\end{table*}

\end{document}